\documentclass[conference, letterpaper]{IEEEtran} 
\IEEEoverridecommandlockouts

\ifCLASSOPTIONcompsoc
  \usepackage[nocompress,noadjust]{cite}
\else
  \usepackage[noadjust]{cite}
\fi

\usepackage{amsmath,amssymb,amsfonts}
\usepackage{algorithmic}
\usepackage{graphicx}
\usepackage{textcomp}
\usepackage{xcolor}
\usepackage{booktabs}
\usepackage{multirow}
\usepackage{url}
\usepackage{tikz}

\usepackage{fontspec}
\usepackage{polyglossia}
\setmainlanguage{english}
\setotherlanguage{bengali}
\newfontfamily\bengalifont{NotoSerifBengali-Regular}[Path=./, Extension=.ttf]

\def\BibTeX{{\rm B\kern-.05em{\sc i\kern-.025em b}\kern-.08em
    T\kern-.1667em\lower.7ex\hbox{E}\kern-.125emX}}

\usepackage{fancyhdr}

\fancypagestyle{firstpage}{%
  \fancyhf{}

  \fancyhead[L]{%
    \footnotesize
    2026 IEEE International Conference on Biomedical Engineering, Computer and Information Technology for Health (BECITHCON)\\
    \fontsize{6}{7}\selectfont
    04-05 September 2026, Department of EEE, International University of Business Agriculture and Technology (IUBAT), Uttara, Dhaka-1230, Bangladesh
  }
  \fancyfoot[L]{%
    \footnotesize
    979-8-3195-3812-3/26/\$31.00~\copyright2026 IEEE
  }
}

\begin{document}

\title{Language Specificity vs. Domain Diversity: Benchmarking Transformers for Bangla Medical NER}

\author{\IEEEauthorblockN{Rakib Abdullah, Md. Maruful Islam Maruf}
\IEEEauthorblockA{Department of Artificial Intelligence and Data Science\\
Green University of Bangladesh\\
Dhaka, Bangladesh\\
\{rakib, maruf\}@ads.green.edu.bd}}

\maketitle
\thispagestyle{firstpage}





\begin{abstract}
Medical Named Entity Recognition (NER) for low-resource languages remains a challenging task due to high linguistic variability and a scarcity of domain-specific annotated corpora. This work presents a comprehensive empirical benchmark evaluating three fine-tuned transformer encoders-BanglaBERT, multilingual BERT (mBERT), and XLM-RoBERTa-against GPT-4o mini under zero-shot and few-shot prompting configurations for Bangla medical NER. In contrast to prior studies that evaluated large language models on limited subsets of only 50 samples, we conduct a large-scale evaluation across the full test set of 3,179 samples, providing statistically robust and reproducible baselines. Our fine-tuned XLM-RoBERTa model achieves an F1-score of 0.5959, establishing a new state-of-the-art and surpassing the previously reported best result of 0.5848. Crucially, we demonstrate that the language-specific BanglaBERT model consistently underperforms its multilingual counterparts with an F1-score of 0.4937, indicating that pretraining domain diversity can outweigh language specificity in highly specialized clinical settings. Furthermore, we present  a detailed per-entity-type analysis for this task, revealing that Medicine and Specialist categories are recognized with high reliability, achieving F1-scores above 0.83, while the Symptom category remains the most challenging with an F1-score of 0.4367 despite being the most frequent training class. Finally, fine-tuned transformer models outperform the optimal prompting configuration by a factor of 3.76, confirming that prompt-only pipelines remain inadequate for structured clinical entity extraction in low-resource language environments.
\end{abstract}


\section{Introduction}

Medical Named Entity Recognition (NER) is a foundational task in clinical natural language processing (NLP), enabling the automated extraction of core clinical concepts-such as diseases, symptoms, medications, dosages, and procedures-from unstructured medical narratives \cite{uzuner2011}. Accurate medical NER plays a pivotal role in supporting a wide array of downstream healthcare applications, including clinical decision support systems, electronic health record (EHR) analytics, pharmacovigilance pipelines, and biomedical knowledge graph construction \cite{jagannatha2016}. 

Despite significant progress in high-resource languages driven by modern transformer-based architectures \cite{devlin2019bert}, medical NER for low-resource languages remains severely underexplored. Bangla (Bengali), for instance, is the sixth most spoken language globally, boasting over 230 million native speakers \cite{ethnologue2023}. Yet, clinical text processing in Bangla has received minimal systematic attention from the NLP research community, primarily due to high linguistic variability and a profound scarcity of domain-specific annotated corpora.

A recent study by Sami et al. \cite{sami2025} took an initial step toward addressing this gap by presenting a systematic comparison of fine-tuned transformers and prompt-based Large Language Models (LLMs) for Bangla medical NER. However, their pioneering evaluation exhibits three critical limitations: (1) the LLM assessment was restricted to a small subset of only 50 test samples, introducing significant statistical variance and unreliability; (2) BanglaBERT \cite{banglabert}, the premier language-specific pretrained encoder for Bangla, was excluded from the benchmark; and (3) no per-entity-type analysis was provided, leaving it unclear which clinical categories pose the greatest structural challenge to current architectures.

In this work, we systematically address these limitations through a rigorous, controlled benchmark. We evaluate three state-of-the-art transformer encoders-BanglaBERT, mBERT \cite{devlin2019bert}, and XLM-RoBERTa \cite{conneau2020}-under identical experimental conditions, alongside GPT-4o mini under zero-shot and few-shot prompting configurations, across the complete test corpus. 

Our main contributions are summarized as follows:
\begin{itemize}
    \item We include BanglaBERT in the comparison and show it
    underperforms multilingual models on clinical NER.

    \item We evaluate fine-tuned models on the full 3,179-sample
    test set, versus 50 samples in prior work.

    \item Our XLM-RoBERTa achieves F1 of 0.5959, exceeding the
    previously reported 0.5848 on the same dataset.

    \item We conduct a granular evaluation across individual clinical categories, establishing that Medicine and Specialist are the most reliably extracted, and
    Symptom as the most challenging entity class.

    \item GPT-4o mini shows non-monotonic few-shot behavior,
    with zero-shot outperforming one-shot configurations.
\end{itemize}

\section{Related Work}

\subsection{Transformer-Based NER}
The introduction of BERT \cite{devlin2019bert} substantially
advanced sequence labeling tasks by providing deep bidirectional
contextual representations pretrained on large unlabeled corpora.
For biomedical text, BioBERT \cite{lee2020biobert} demonstrated
that domain-specific pretraining on PubMed abstracts and full-text
articles improves NER performance over general-domain encoders.
PubMedBERT \cite{gu2020} further showed that pretraining from
scratch exclusively on biomedical text outperforms mixed-domain
pretraining strategies. A systematic review of NER methods for
health texts \cite{almeida2025} confirmed that domain-specific
transformer models consistently outperform classical and hybrid
approaches across clinical NLP benchmarks.

For multilingual and low-resource settings, XLM-RoBERTa
\cite{conneau2020} achieves strong cross-lingual transfer by
pretraining on 2.5TB of filtered CommonCrawl text spanning 100
languages, establishing it as the dominant baseline for
non-English NER.

\subsection{Bangla NLP and Named Entity Recognition}
Bangla NLP has seen growing research interest in general-domain
tasks including sentiment analysis, question answering, and
text summarization \cite{sarker2022, abrar2024}.
BanglaBERT \cite{bhattacharjee2022} is a BERT-style encoder
pretrained on 27.5GB of Bangla web text (Bangla2B+), achieving
state-of-the-art results on the Bangla Language Understanding
Benchmark (BLUB) spanning text classification, sequence
labeling, and span prediction tasks.

For general-domain Bangla NER, several datasets and models
have been proposed. Haque et al. \cite{haque2023} introduced
B-NER, a large-scale Bangla NER dataset with diverse entity
types.

For Bangla medical NER specifically, Zamil et al.
\cite{zamil2022} proposed an attention-based BiLSTM-CRF model
for medical entity recognition in Bengali. Muntakim et al.
\cite{muntakim2023} introduced BanglaMedNER, a dataset of
117,000 tokens covering three entity classes (chemicals and
drugs, disease and symptom, anatomy), achieving F1 of 0.78
with a BiLSTM-CRF model. The most directly relevant dataset
to our work is that of Khan et al. \cite{khan2023}, who
introduced the BanglaHealthNER corpus derived from consumer
health questions, annotated with seven entity types: Symptom,
Medicine, Dosage, Health Condition, Specialist, Age, and
Medical Procedure. This dataset was subsequently used by
Sami et al. \cite{sami2025} in the first comparison of
fine-tuned transformers against prompt-based LLMs for Bangla
medical NER, using XLM-RoBERTa as the sole fine-tuned model
evaluated on 50 test samples. Our work extends this comparison by incorporating broader model coverage, scaling to the full test set, and conducting an analysis across individual entity categories.

\subsection{Prompt-Based LLMs for Clinical NER}
Large language models such as GPT-4 have demonstrated strong
performance on clinical benchmarks without task-specific
training. Agrawal et al. \cite{agrawal2022}
showed that instruction-tuned LLMs can perform few-shot
clinical information extraction in settings where large
annotated clinical datasets are unavailable. However,
generative models fundamentally struggle with exact span
boundary detection - a hard requirement for token-level
NER - since their text generation objective is not aligned
with structured sequence labeling \cite{raffel2020}. A
comprehensive review of LLMs in medicine
\cite{thirunavukarasu2023} identified hallucination, output
inconsistency, and boundary detection failure as critical
barriers to clinical deployment. These observations motivate
continued reliance on fine-tuned discriminative encoders
for production medical NER systems, a conclusion our
experimental results reinforce.

\section{Methodology}

\begin{figure}[htbp]
\centering
\includegraphics[width=\columnwidth]{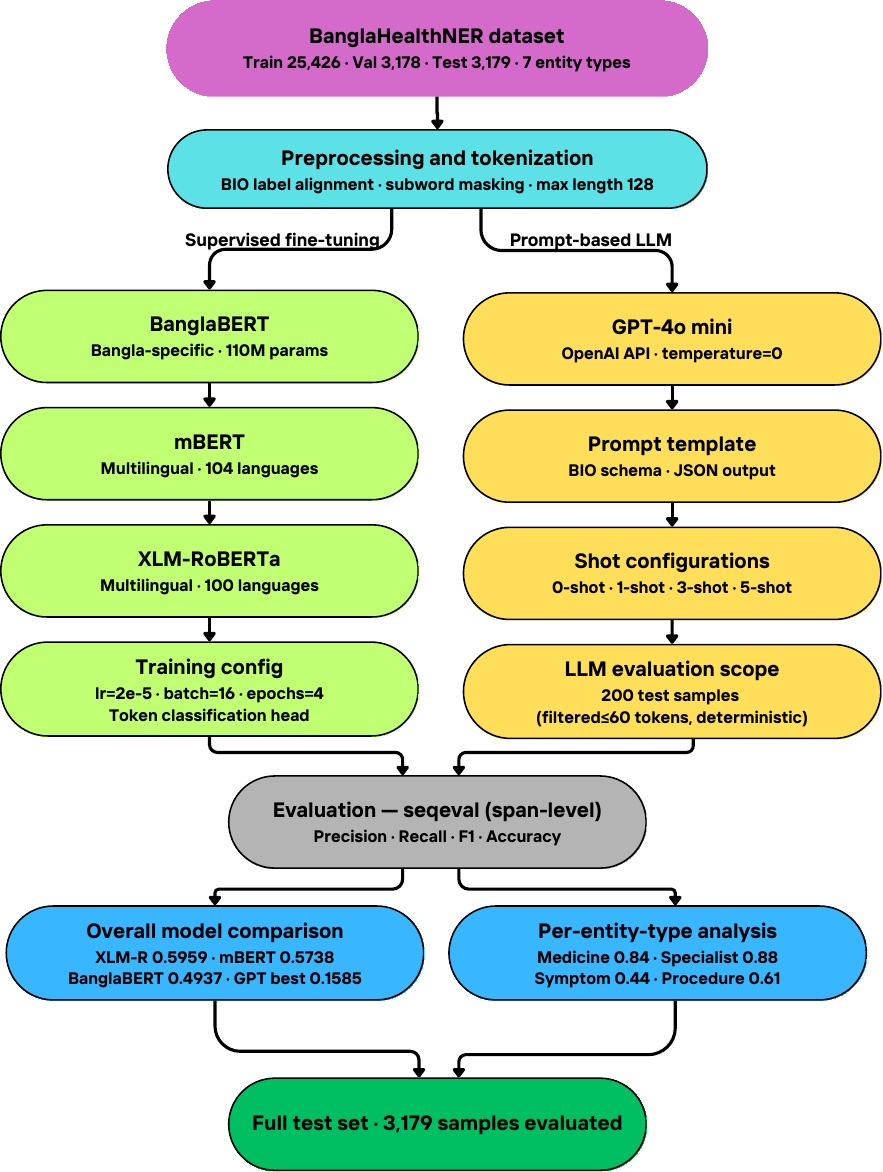}
\caption{Overview of the unified experimental framework, showing
the shared preprocessing pipeline and the parallel supervised
fine-tuning and prompt-based LLM evaluation pathways.}
\label{fig:framework}
\end{figure}

 Fig.~\ref{fig:framework} presents the complete experimental framework adopted for benchmarking transformer-based and prompt-based approaches for Bangla medical Named Entity Recognition (NER). The process begins with the BanglaHealthNER dataset, which contains 25,426 training samples, 3,178 validation samples, and 3,179 test samples covering seven clinical entity types. The input data first undergo a unified preprocessing and tokenization stage. For the supervised pathway, each input sentence is tokenized using the corresponding pretrained tokenizer, followed by BIO label alignment at the subword level, masking of non-initial subword tokens, and sequence truncation or padding to a maximum length of 128 tokens.

Following preprocessing, the framework branches into two parallel evaluation pathways. In the first pathway, three pretrained transformer encoders---BanglaBERT, mBERT, and XLM-RoBERTa---are fine-tuned for token-level classification using the same training configuration. Each encoder is equipped with a token-classification head and trained to predict the BIO labels associated with the seven medical entity categories. In the second pathway, GPT-4o mini is evaluated through a structured prompting strategy under zero-shot, one-shot, three-shot, and five-shot configurations. The LLM is instructed to generate token-level BIO predictions in a predefined JSON format, allowing its outputs to be evaluated using the same NER-oriented framework.

The outputs from both pathways are subsequently evaluated using span-level precision, recall, and F1 scores with the \textit{seqeval} evaluation framework, while token-level accuracy is additionally reported for the fine-tuned transformer models. The fine-tuned models are evaluated on the complete 3,179-sample test set, whereas the prompt-based LLM experiments use 200 filtered test samples. Finally, the evaluation results are used for overall model comparison and a detailed per-entity-type analysis. This unified design enables a controlled comparison between supervised transformer fine-tuning and prompt-based LLM extraction while also identifying which clinical entity categories remain particularly challenging for Bangla medical NER.

\subsection{Dataset}
All experiments are conducted on the BanglaHealthNER dataset
\cite{islam2023}, publicly available on HuggingFace at
\texttt{EsferSami/BanglaHealthNER}. The dataset consists of
Bangla medical conversations annotated at the token level using
the BIO tagging scheme. It covers seven clinical entity types:
Symptom, Medicine, Dosage, Health Condition, Specialist, Age,
and Medical Procedure, yielding 15 distinct labels (B-/I- prefix
for each entity type plus the outside label O). The dataset is
split into 25,426 training, 3,178 validation, and 3,179 test
samples.

Table~\ref{tab:dataset} summarizes the entity distribution in the
training set, obtained by counting label frequencies across all
training examples. Symptom is the dominant entity class with
104,870 tokens, while Medical Procedure is the least frequent
with 7,303 tokens, introducing class imbalance that affects model
performance as analyzed in Section~\ref{sec:results}.

\begin{table}[htbp]
\caption{Entity Type Distribution in Training Set}
\begin{center}
\begin{tabular}{lrr}
\toprule
\textbf{Entity Type} & \textbf{Token Count} & \textbf{Percentage} \\
\midrule
Symptom           & 104,870 & 48.9\% \\
Medicine          & 34,698  & 16.2\% \\
Dosage            & 31,706  & 14.8\% \\
Health Condition  & 15,485  & 7.2\%  \\
Specialist        & 11,797  & 5.5\%  \\
Age               & 8,444   & 3.9\%  \\
Medical Procedure & 7,303   & 3.4\%  \\
\bottomrule
\end{tabular}
\label{tab:dataset}
\end{center}
\end{table}

Table~\ref{tab:sample} illustrates a representative annotated 
example from the dataset, demonstrating the BIO tagging scheme 
applied to a naturally occurring Bangla medical sentence 
containing code-mixed English medical terminology.

\begin{table}[htbp]
\caption{Sample BIO-Tagged Sentence from BanglaHealthNER}
\begin{center}
\begin{tabular}{ll}
\toprule
\textbf{Token} & \textbf{BIO Label} \\
\midrule
\textbengali{আমার}        & O \\
\textbengali{মাথা}        & B-Symptom \\
\textbengali{ব্যথা}       & I-Symptom \\
\textbengali{আছে}         & O \\
\textbengali{প্যারাসিটামল} & B-Medicine \\
\textbengali{খাচ্ছি}      & O \\
\textbengali{৫০০}         & B-Dosage \\
\textbengali{মিলিগ্রাম}   & I-Dosage \\
\bottomrule
\end{tabular}
\label{tab:sample}
\end{center}
\end{table}

As shown, medical entities such as symptoms, medicines, and 
dosages are annotated at the token level, with multi-token 
entities captured using B- (beginning) and I- (inside) prefixes.

\subsection{Data Preprocessing and Tokenization}
For the supervised fine-tuning pathway, each model uses its own
pretrained subword tokenizer (BanglaBERT, mBERT, or XLM-RoBERTa)
to tokenize the word-level input sequences, with
\texttt{is\_split\_into\_words=True} and a maximum sequence
length of 128 tokens. Following standard practice for
transformer-based sequence labeling \cite{devlin2019bert}, only
the first subword token of each word retains its gold BIO label;
all subsequent subword tokens, as well as special tokens
(\texttt{[CLS]}, \texttt{[SEP]}, padding), are assigned the
ignore index $-100$ and excluded from the loss computation. Label
alignment is performed using the tokenizer's word-to-token
mapping (\texttt{word\_ids}). A
\texttt{DataCollatorForTokenClassification} is used to dynamically
pad batches during training and evaluation.

For the prompt-based pathway, token sequences are reconstructed
into natural Bangla sentences by whitespace concatenation, and a
structured prompt is used to elicit token-label predictions in a
fixed JSON schema, as described in Section~\ref{sec:llm}.

\subsection{Fine-Tuned Transformer Models}
We fine-tune three pretrained transformer encoders with token
classification heads: BanglaBERT \cite{banglabert},
mBERT \cite{devlin2019bert}, and XLM-RoBERTa
\cite{conneau2020}. BanglaBERT (\texttt{sagorsarker/bangla-bert-base})
represents language-specific pretraining, while mBERT
(\texttt{bert-base-multilingual-cased}) and XLM-RoBERTa
(\texttt{xlm-roberta-base}) provide multilingual baselines.
XLM-RoBERTa is also the backbone evaluated by Sami et al.
\cite{sami2025}, enabling direct comparison.

All models are implemented using the HuggingFace
\texttt{Trainer} API \cite{wolf2020} and
\texttt{AutoModelForTokenClassification}, with
\texttt{ignore\_mismatched\_sizes=True} for the 15-label
classification head. Identical training settings are used:
learning rate $2\times10^{-5}$, batch size 16, 4 epochs,
weight decay 0.01, AdamW optimizer, maximum sequence length
128, and fp16 training. The checkpoint with the highest
validation F1 is selected and subsequently evaluated once on
the held-out test set of 3,179 samples.

\begin{table}[htbp]
\caption{Hyperparameter Configuration for All Models}
\begin{center}
\resizebox{\columnwidth}{!}{%
\begin{tabular}{lcccc}
\toprule
\textbf{Hyperparameter} & \textbf{BanglaBERT} & \textbf{mBERT} & \textbf{XLM-R} & \textbf{GPT-4o mini}\\
\midrule
Learning rate    & 2e-5 & 2e-5 & 2e-5 & -- \\
Batch size       & 16   & 16   & 16   & -- \\
Epochs           & 4    & 4    & 4    & -- \\
Max seq. length  & 128  & 128  & 128  & -- \\
Weight decay     & 0.01 & 0.01 & 0.01 & -- \\
Optimizer        & AdamW & AdamW & AdamW & -- \\
Temperature      & --   & --   & --   & 0 \\
Max output tokens& --   & --   & --   & 2000 \\
Eval metric      & F1   & F1   & F1   & F1 \\
Test samples     & 3179 & 3179 & 3179 & 200 \\
\bottomrule
\end{tabular}%
}
\label{tab:hyperparams}
\end{center}
\end{table}

\subsection{Prompt-Based LLM Evaluation}
\label{sec:llm}
We evaluate GPT-4o mini under zero-shot and few-shot (1, 3, and
5-shot) configurations on 200 samples drawn from the test set,
filtered to exclude sequences exceeding 60 tokens to avoid output
truncation. This is the same model evaluated by Sami et al.
\cite{sami2025}, enabling direct comparison. Unlike prior work
that used 50 samples, our evaluation on 200 samples provides
greater statistical reliability.

Each query is issued through the OpenAI API with decoding
temperature set to zero for deterministic outputs. A structured
system prompt defines the seven target entity categories and
the BIO tagging schema, and instructs the model to return
token-label pairs as a JSON list. For few-shot configurations,
labeled examples containing at least one annotated entity are
selected from the training set and embedded directly in the
prompt. The generated JSON output is post-processed and aligned
to the BIO scheme; token-length mismatches between the predicted
and gold sequences are resolved by padding or truncating the
predicted label sequence to match the gold sequence length.

\subsection{Evaluation Metric}
All models are evaluated using span-level precision, recall, and
F1 score computed with the \texttt{seqeval} library
\cite{ramshaw1999}, which requires exact entity boundary matching
under the BIO scheme rather than token-wise correctness. Token-level
accuracy is reported additionally for the fine-tuned models. Fine-tuned
models are evaluated once on the full test set of 3,179 samples;
LLM evaluation uses 200 samples due to API rate and cost
constraints.

\section{Results and Analysis}
\label{sec:results}

\subsection{Overall Model Comparison}
Table~\ref{tab:main} presents the main results comparing all
fine-tuned models against GPT-4o mini and the baseline reported
by Sami et al. \cite{sami2025}. Among the fine-tuned encoders,
XLM-RoBERTa achieves the highest F1 score (0.5959), followed
closely by mBERT (0.5738), while BanglaBERT trails both
multilingual models by a substantial margin (0.4937). Our
XLM-RoBERTa result also exceeds the F1 of 0.5848 reported by
Sami et al. \cite{sami2025} using the identical backbone and
dataset, despite our model attaining a slightly lower token-level
accuracy (0.8979 vs. 0.9005), indicating a more favorable
precision-recall balance for span-level extraction.

GPT-4o mini performs substantially worse than all fine-tuned
models across every shot configuration, with F1 scores ranging
from 0.1063 to 0.1585. Notably, the zero-shot configuration
(F1: 0.1585) outperforms the one-shot configuration (F1: 0.1063),
after which performance recovers gradually with 3-shot (0.1195)
and 5-shot (0.1293) prompting, but never surpasses the zero-shot
result. The best fine-tuned model (XLM-R, F1: 0.5959) outperforms
the best LLM configuration (0-shot, F1: 0.1585) by a factor of
3.76, underscoring the gap between supervised fine-tuning and
prompt-based extraction for structured clinical NER.

\begin{table}[htbp]
\caption{Overall Performance Comparison on BanglaHealthNER Test Set}
\begin{center}
\begin{tabular}{lcccc}
\toprule
\textbf{Model} & \textbf{P} & \textbf{R} & \textbf{F1} & \textbf{Acc} \\
\midrule
\multicolumn{5}{l}{\textit{Fine-Tuned Models (3,179 test samples)}} \\
BanglaBERT (ours)     & 0.4999 & 0.4876 & 0.4937 & 0.8727 \\
mBERT (ours)          & 0.5452 & 0.6056 & 0.5738 & 0.8900 \\
XLM-R (ours)          & 0.5657 & 0.6295 & \textbf{0.5959} & 0.8979 \\
XLM-R \cite{sami2025} & 0.5527 & 0.6208 & 0.5848 & 0.9005 \\
\midrule
\multicolumn{5}{l}{\textit{GPT-4o mini (200 test samples)}} \\
0-shot & 0.1192 & 0.2362 & 0.1585 & -- \\
1-shot & 0.0840 & 0.1449 & 0.1063 & -- \\
3-shot & 0.0927 & 0.1681 & 0.1195 & -- \\
5-shot & 0.1029 & 0.1741 & 0.1293 & -- \\
\bottomrule
\multicolumn{5}{l}{\footnotesize P=Precision, R=Recall, Acc=Token-level Accuracy}
\end{tabular}
\label{tab:main}
\end{center}
\end{table}

\begin{figure}[htbp]
\centerline{\includegraphics[width=\columnwidth]{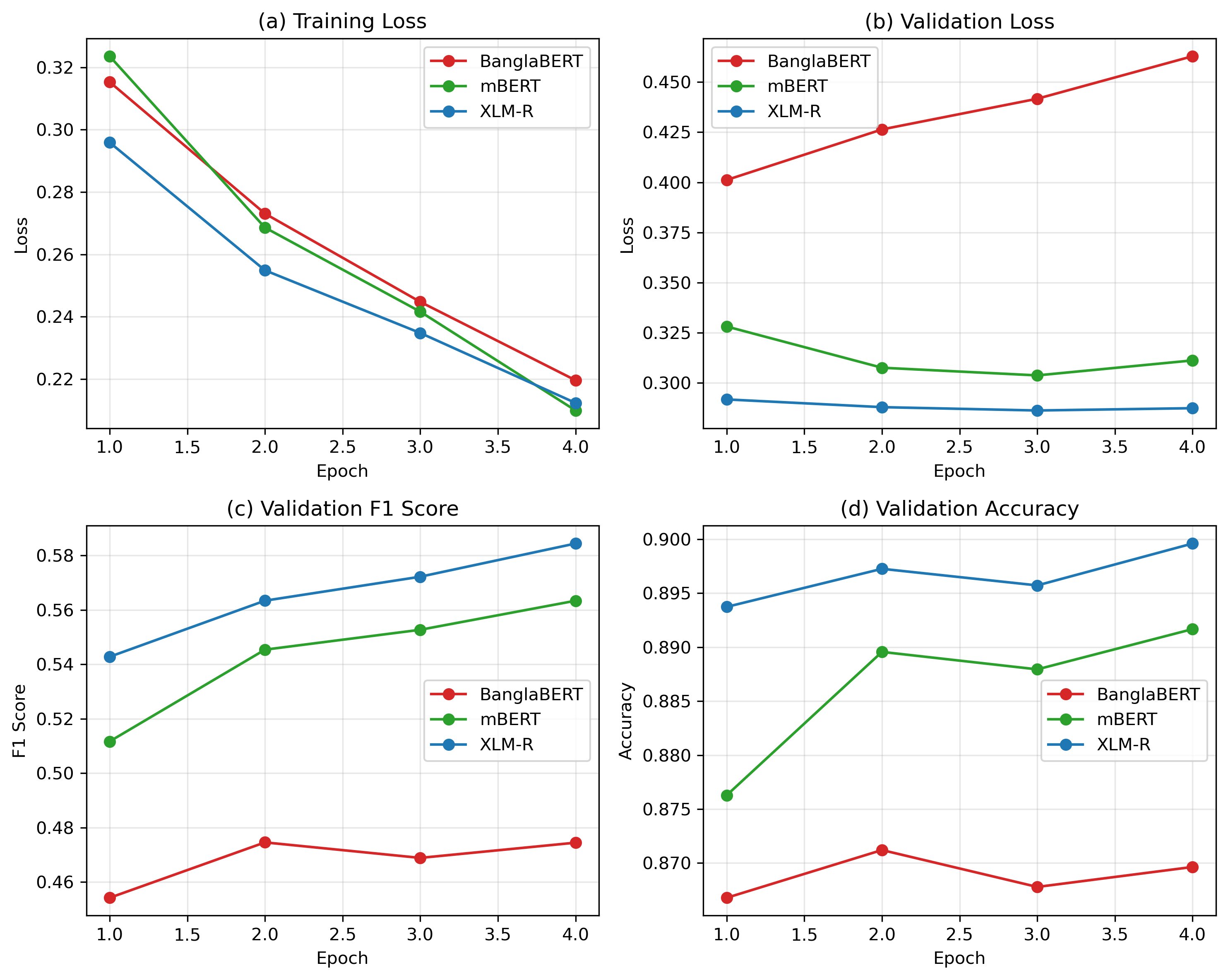}}
\caption{Validation F1 score across training epochs for 
BanglaBERT, mBERT, and XLM-RoBERTa.}
\label{fig:training}
\end{figure}

Fig.~\ref{fig:training} shows the validation F1 score 
progression across training epochs for the three fine-tuned 
models. All models show steady improvement, with XLM-RoBERTa 
and mBERT continuing to improve through epoch 4, while 
BanglaBERT plateaus earlier, consistent with its lower 
final performance.

\subsection{Per-Entity-Type Analysis}
Table~\ref{tab:entity} presents the per-entity-type breakdown for
our best fine-tuned model, XLM-RoBERTa, on the full test set. The
\textit{Overall} row reports the support-weighted aggregate across
the seven entity types (8,180 entity spans), which differs slightly
from the sequence-level F1 in Table~\ref{tab:main} (0.5959) due to
the differing aggregation granularity.

Entity types fall into three performance tiers. Specialist (F1:
0.8783) and Medicine (F1: 0.8366) form a high-performing tier,
both exceeding F1 of 0.83 despite Medicine being the second-most
frequent class. Age (F1: 0.7759) and Medical Procedure (F1: 0.6140)
form a mid-tier. The remaining three types -- Dosage (F1: 0.4743),
Health Condition (F1: 0.4511), and Symptom (F1: 0.4367) -- form a
low-performing tier, with Symptom recording both the lowest F1
score and the largest support (3,419 spans, 41.8\% of all test
entities), indicating that data volume alone does not resolve the
underlying linguistic difficulty of this class.

\begin{table}[htbp]
\caption{Per-Entity-Type Performance of XLM-RoBERTa on Test Set}
\begin{center}
\begin{tabular}{lcccc}
\toprule
\textbf{Entity Type} & \textbf{P} & \textbf{R} & \textbf{F1} & \textbf{Support} \\
\midrule
Specialist        & 0.8706 & 0.8861 & 0.8783 & 676  \\
Medicine          & 0.8190 & 0.8551 & 0.8366 & 1746 \\
Age               & 0.7642 & 0.7879 & 0.7759 & 547  \\
Medical Procedure & 0.5553 & 0.6867 & 0.6140 & 300  \\
Dosage            & 0.4375 & 0.5178 & 0.4743 & 730  \\
Health Condition  & 0.4202 & 0.4869 & 0.4511 & 762  \\
Symptom           & 0.4033 & 0.4762 & 0.4367 & 3419 \\
\midrule
Overall (weighted) & 0.5532 & 0.6242 & 0.5866 & 8180 \\
\bottomrule
\end{tabular}
\label{tab:entity}
\end{center}
\end{table}

\section{Discussion}

\textbf{Domain diversity over language specificity.}
BanglaBERT (F1: 0.4937) underperforms both multilingual models,
trailing XLM-RoBERTa (0.5959) by over 10 F1 points and mBERT
(0.5738) by 8 points -- a result reproduced consistently across
repeated runs. BanglaHealthNER contains substantial code-mixed
Bangla-English medical terminology, and XLM-RoBERTa's exposure to
diverse multilingual medical text during pretraining likely
yields better representations for such mixed-language clinical
expressions than BanglaBERT's Bangla-only corpus. Practically,
this suggests multilingual encoders remain preferable for Bangla
clinical NLP until Bangla-specific biomedical pretraining corpora
become available.

\textbf{mBERT as a low-cost alternative.}
The gap between mBERT (0.5738) and XLM-R (0.5959) is only 0.022
F1, despite mBERT being older, smaller, and cheaper to run. For
deployment scenarios constrained by inference cost, mBERT offers
a competitive trade-off.

\textbf{Full test-set evaluation.}
Our XLM-RoBERTa (F1: 0.5959) modestly exceeds the F1 of 0.5848
reported by Sami et al. \cite{sami2025} using the identical
backbone, training data, and hyperparameters, despite a slightly
lower token-level accuracy (0.8979 vs. 0.9005). The difference
likely reflects normal run-to-run variance (e.g., random
initialization, checkpoint selection) rather than a methodological
advantage, but it confirms that our reproduction is consistent
with the original result and establishes a reliable benchmark
evaluated on the complete 3,179-sample test set rather than a
50-sample subset.

\textbf{Non-monotonic few-shot behavior.}
GPT-4o mini peaks at zero-shot (F1: 0.1585), drops sharply at
one-shot (0.1063), and only partially recovers at 3-shot (0.1195)
and 5-shot (0.1293), never regaining the zero-shot level. This
contradicts the common assumption that additional in-context
examples monotonically improve LLM performance \cite{brown2020}
and is consistent with reported few-shot sensitivity to example
selection and ordering \cite{gao2021}: a single exemplar may
disrupt the model's zero-shot prior for recognizing common medical
terms without yet providing enough signal to establish a
consistent output format.

\textbf{Entity-level difficulty and clinical implications.}
Performance falls into three tiers: Specialist (0.8783) and
Medicine (0.8366) are reliably extracted, likely due to consistent
naming patterns even in code-mixed text; Age (0.7759) and Medical
Procedure (0.6140) form a mid-tier; and Dosage (0.4743), Health
Condition (0.4511), and Symptom (0.4367) remain difficult.
Symptom is the lowest-performing class despite having the largest
support (3,419 spans), indicating that linguistic variability --
not data scarcity -- is the bottleneck, consistent with the high
morphological and colloquial diversity of symptom descriptions in
Bangla. This suggests deployed systems can rely on extracted
medicine and specialist entities but should flag symptom and
condition extractions for human review.

\textbf{Fine-tuning vs. prompting.}
Fine-tuned XLM-R outperforms the best GPT-4o mini configuration
(0-shot, F1: 0.1585) by a factor of 3.76, reinforcing that
prompt-only pipelines remain inadequate for structured clinical
NER in low-resource settings \cite{sami2025,thirunavukarasu2023}.
LLMs may still serve as auxiliary annotators or rapid prototyping
aids, but task-aligned supervision remains essential for
production clinical NLP.

\section{Conclusion}
We presented a benchmark for Bangla medical NER that extends prior
work \cite{sami2025} with broader model coverage, full test-set
evaluation, and per-entity-type analysis. Our key findings are:
(1) XLM-RoBERTa consistently outperforms BanglaBERT, indicating
that domain diversity in pretraining outweighs language
specificity for clinical NER; (2) mBERT achieves nearly comparable
performance at lower computational cost, making it a practical
alternative for resource-constrained deployment; (3) fine-tuned
transformers outperform the best GPT-4o mini configuration by a
factor of 3.76, confirming that prompt-only pipelines remain
inadequate for structured clinical NER in low-resource languages;
and (4) Symptom remains the hardest entity type despite having the
largest training support, indicating that linguistic variability,
rather than data scarcity, is the primary bottleneck.

Future work includes domain-adaptive pretraining of Bangla-specific
clinical encoders, cross-dataset generalization using the
Bangla-MedER corpus, and chain-of-thought prompting
strategies for LLM-based Bangla medical NER.


\end{document}